\documentclass[letterpaper, 10 pt, conference]{ieeeconf}  

\IEEEoverridecommandlockouts                              

\usepackage{algorithm}
\usepackage{algorithmic}

\usepackage{booktabs}
\usepackage{amsmath}
\usepackage{amssymb}
\usepackage{float}
\usepackage{graphicx}
\usepackage{listings}
\usepackage{multirow}
\usepackage[table,xcdraw]{xcolor}
\usepackage{adjustbox}
\usepackage{makecell}
\usepackage{graphicx}
\usepackage{subcaption}

\title{\LARGE \bf

IMU-Enhanced EEG Motion Artifact Removal with Fine-Tuned Large Brain Models 

}

\author{ Yuhong Zhang$^{1\dag}$, Xusheng Zhu$^{2 \dag}$, Yuchen Xu$^{1}$, ChiaEn Lu$^{2}$, Hsinyu Shih$^{1}$, \\ Gert Cauwenberghs$^{1}$, Tzyy-Ping Jung$^{1}$ 
\thanks{$\dag$ Equal Contribution} 
\thanks{$^{1}$ Shu Chien-Gene Lay Department of Bioengineering and Institute for Neural Computation, University of California, San Diego (UCSD), La Jolla, CA 92093, USA. {\tt\small yuz291@ucsd.edu, yux013@ucsd.edu, tpjung@ucsd.edu, gcauwenberghs@ucsd.edu}}%
\thanks{$^{2}$ Department of Computer Science and Engineering, University of California, San Diego (UCSD), La Jolla, CA 92093, USA. {\tt\small xuz056@ucsd.edu, chl298@ucsd.edu }}%
}

\begin{document}

\maketitle
\thispagestyle{empty}
\pagestyle{empty}

\begin{abstract}
Electroencephalography (EEG) is a non-invasive method for measuring brain activity with high temporal resolution; however, EEG signals often exhibit low signal-to-noise ratios because of contamination from physiological and environmental artifacts. One of the major challenges hindering the real-world deployment of brain-computer interfaces (BCIs) involves the frequent occurrence of motion-related EEG artifacts.
Most prior studies on EEG motion artifact removal rely on single-modality approaches, such as Artifact Subspace Reconstruction (ASR) and Independent Component Analysis (ICA), without incorporating simultaneously recorded modalities like inertial measurement units (IMUs), which directly capture the extent and dynamics of motion.
This work proposes a fine-tuned large brain model (LaBraM)-based correlation attention mapping method that leverages spatial channel relationships in IMU data to identify motion-related artifacts in EEG signals. The fine-tuned model contains approximately 9.2 million parameters and uses 5.9 hours of EEG and IMU recordings for training, just 0.2346\% of the 2500 hours used to train the base model.
We compare our results against the established ASR-ICA benchmark across varying time scales and motion activities, showing that incorporating IMU reference signals significantly improves robustness under diverse motion scenarios.
\\


\end{abstract}


\section{INTRODUCTION}
EEG is a widely used non-invasive technique for recording brain activity, valued for its high temporal resolution and versatility across both clinical and research domains. Despite these advantages, EEG signals often suffer a low signal-to-noise ratio, primarily because of contamination from physiological sources such as eye blinks, muscle activity,  cardiac rhythms, and environmental factors like electrical interference \cite{mullen2013real}.

Over the years, researchers have developed a range of signal processing techniques to mitigate these artifacts, with Independent Component Analysis (ICA), Artifact Subspace Reconstruction (ASR), and its extensions emerging as foundational tools \cite{chang2018evaluation, chang2019evaluation}. ICA-based methods have proven effective in isolating and removing non-neural components, forming the backbone of many artifact removal pipelines \cite{jung2000removing,chuang2022ic}. 

However, in real-world brain-computer interface (BCI) applications, users are no longer stationary. Instead, they are expected to engage in natural behaviors that involve frequent and intensive daily physical movement, like walking, running, and even professional sports. This poses a significant challenge, as motion-related artifacts can severely degrade EEG signal quality. Even subtle movements, like slight head tilts, can introduce substantial noise, making the signal highly vulnerable.

Inertial measurement units (IMUs) are compact sensor devices commonly embedded in consumer electronics and wearable systems to measure acceleration, angular velocity, and orientation. Leveraging IMUs as reference signals for EEG artifact removal is intuitive, as these sensors directly quantify the intensity and dynamics of motion that often corrupt EEG recordings. Several studies have demonstrated effective IMU-based EEG artifact suppression. For instance, Kilicarslan et al. used adaptive filtering with IMU-derived signals to remove gait-related artifacts during walking tasks \cite{kilicarslan2019characterization}. Similarly, Beach et al. embedded IMUs directly into EEG electrodes and applied normalized least-mean-square adaptive filters, substantially reducing EEG artifacts resulting from vigorous head movements \cite{beach2021motion}. Lee et al. use IMU channels to define the artifact subspace in ICA and apply an adaptive Recursive Least Squares (RLS) algorithm to continuously update and remove artifact estimates from the EEG in real time \cite{lee2020real}.

Recent approaches further leverage IMU data using advanced machine learning and sensor fusion techniques. Downey et al. developed the iCanClean algorithm, employing canonical correlation analysis (CCA) with IMU-derived reference signals to significantly enhance EEG quality across diverse artifact conditions without requiring extensive calibration \cite{downey2023icanclean}. While these IMU-assisted methods consistently improve EEG artifact removal performance, they typically share limitations, such as the need for precise synchronization between EEG and IMU data streams and limited generalizability beyond specific experimental scenarios or movement patterns. Additionally, few studies have explored using large-scale deep neural network models to integrate IMU data directly for EEG motion artifact removal.

Current state-of-the-art deep learning approaches rely on large models that pretrain a core backbone on massive amounts of domain-specific generic data and then fine-tune it for downstream tasks. This study proposes fine-tuning pretrained large brain model decoders by integrating IMU signals as reference modalities to remove EEG motion artifacts, bridging the gap between conventional digital filtering and advanced deep learning approaches.
We benchmark our method against the widely adopted ASR+ICA pipeline, evaluating performance across varying timescales and motion activities to demonstrate its robustness.

The remainder of this study is organized as follows: Section II introduces the EEG-IMU motion dataset, Section III details our proposed motion artifact removal methodology, Section IV presents the evaluation results, and Section V concludes with discussions and future directions.

\section{Dataset}

\begin{figure}[t]
    \centering
    \includegraphics[width=0.7\linewidth]{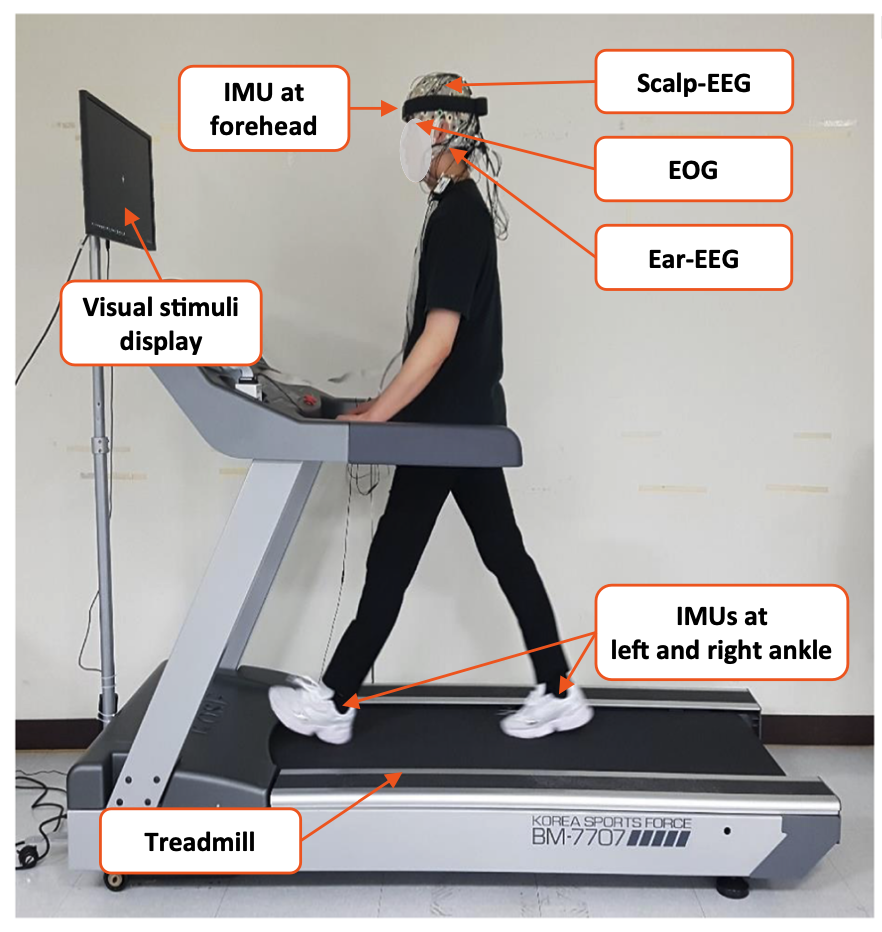}
    \caption{The experimental setup used in the dataset \cite{lee2021mobile}. For the primary goal of motion artifact removal, we used IMU data from the forehead along with scalp EEG recordings. }
    \label{fig:dataset_overview}
\end{figure}

In this study, we use the Mobile BCI dataset by Lee et al. \cite{lee2021mobile}, which includes scalp and ear EEG recordings collected during Event-Related Potential (ERP) and Steady-State Visual Evoked Potential (SSVEP) paradigms under various movement conditions, including standing, walking, and running, see Fig. \ref{fig:dataset_overview}.

This dataset includes two types of BCI paradigms to evaluate EEG signal quality in mobile environments. In the ERP task, participants identified target (`OOO') and non-target (`XXX') stimuli, each shown for 0.5 seconds across 300 trials. In the SSVEP task, participants focused on one of three flickering stimuli displayed at 5.45, 8.57, and 12 Hz at the left, center, and right positions, respectively. Each trial included 2 seconds of cueing, 5 seconds of flicker, and 2 seconds of rest, with 60 trials in total.

Twenty-four participants took part in both tasks, each contributing approximately 72 minutes of data, with ERP and SSVEP sessions recorded under four movement conditions: standing, slow walking (0.8 m/s), fast walking (1.6 m/s), and slight running (2.0 m/s), each lasting 7–8 minutes. For this study, we selected ERP recordings from 11 participants under the three active conditions (slow walking, fast walking, and slight running) for model training, using the standing condition as a benchmark. Each participant contributed around 32 minutes of usable data, yielding a total of approximately 5.9 hours. Participants with missing IMU or EEG segments were excluded from the current analysis.

EEG was recorded using 32 Ag/AgCl electrodes placed according to the international 10/20 system with the BrainAmp system (Brain Products GmbH). Head motion was captured using a 9-axis IMU (APDM wearable technologies), including a 3-axis accelerometer, gyroscope, and magnetometer, sampled at 128 Hz with 32-bit resolution. We focused only on the head-mounted EEG and IMU recordings, excluding ear-EEG and additional body-worn sensors.

Prior to model input, EEG signals underwent preprocessing, including removal of unused channels, bandpass filtering from 0.1 to 75 Hz, 60 Hz notch filtering, resampling to 200 Hz, and unit conversion to microvolts (µV).

\section{Method}

\begin{figure*}[t]
    \centering
    \includegraphics[width=\linewidth]{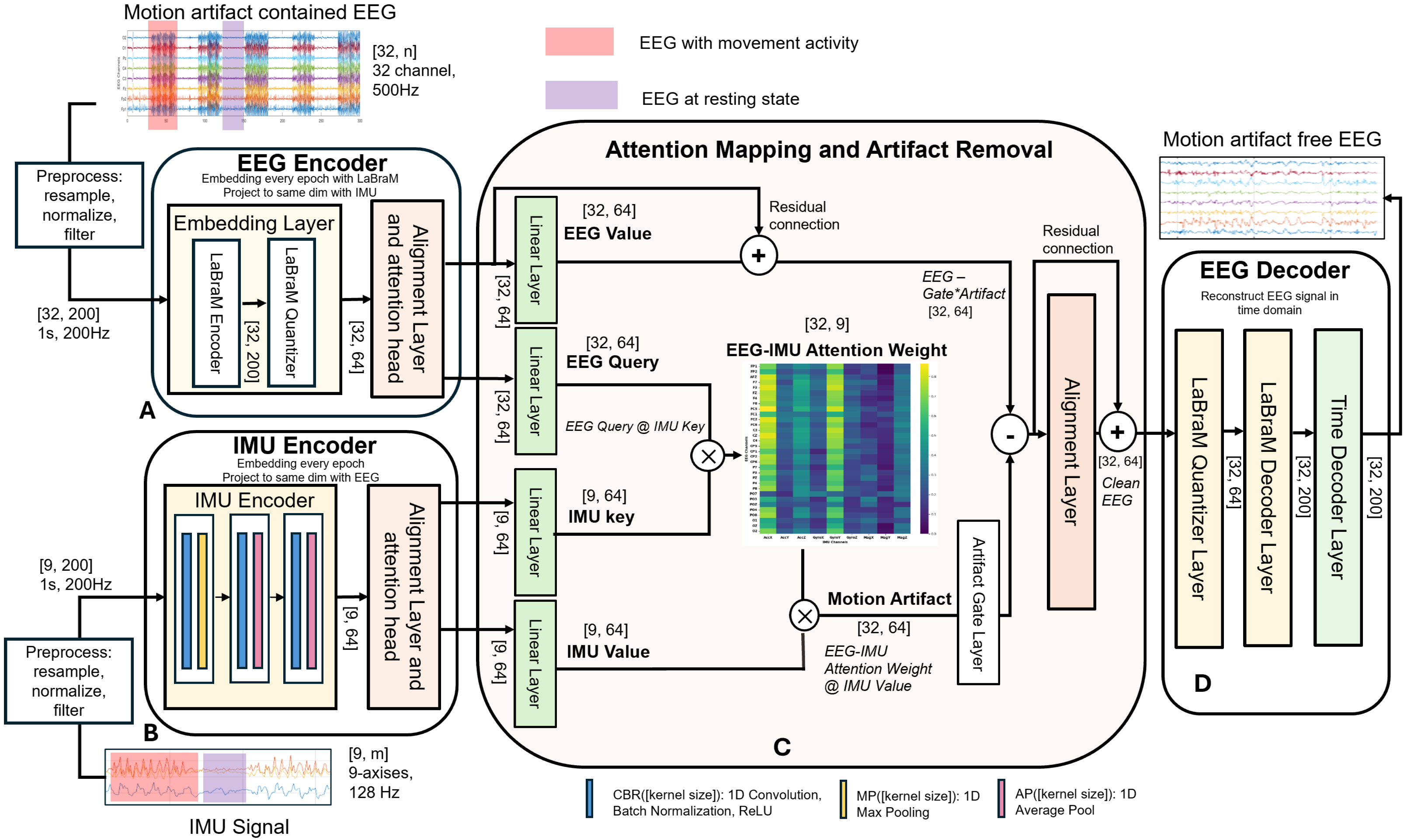}
    \caption{Overview of the proposed IMU-integrated motion artifact removal framework, which fine-tunes a LaBraM-based model. The system comprises four main modules: (A) an EEG encoder that extracts spatiotemporal representations from motion-contaminated EEG using a pretrained LaBraM encoder and quantizer; (B) an IMU encoder that embeds IMU signals into the same dimension as EEG features; (C) an attention mapping module that computes EEG–IMU correlations using attention weights to identify motion artifacts, then subtracts them to recover clean EEG in a low-dimensional embedding space; and (D) an EEG decoder that reconstructs artifact-free EEG in the time domain.}
    \label{fig2}
\end{figure*}

\subsection{Fine-tune Large Brain Model using Attention Mapping}

\begin{figure*}[htbp]
    \centering
    \includegraphics[width=\linewidth]{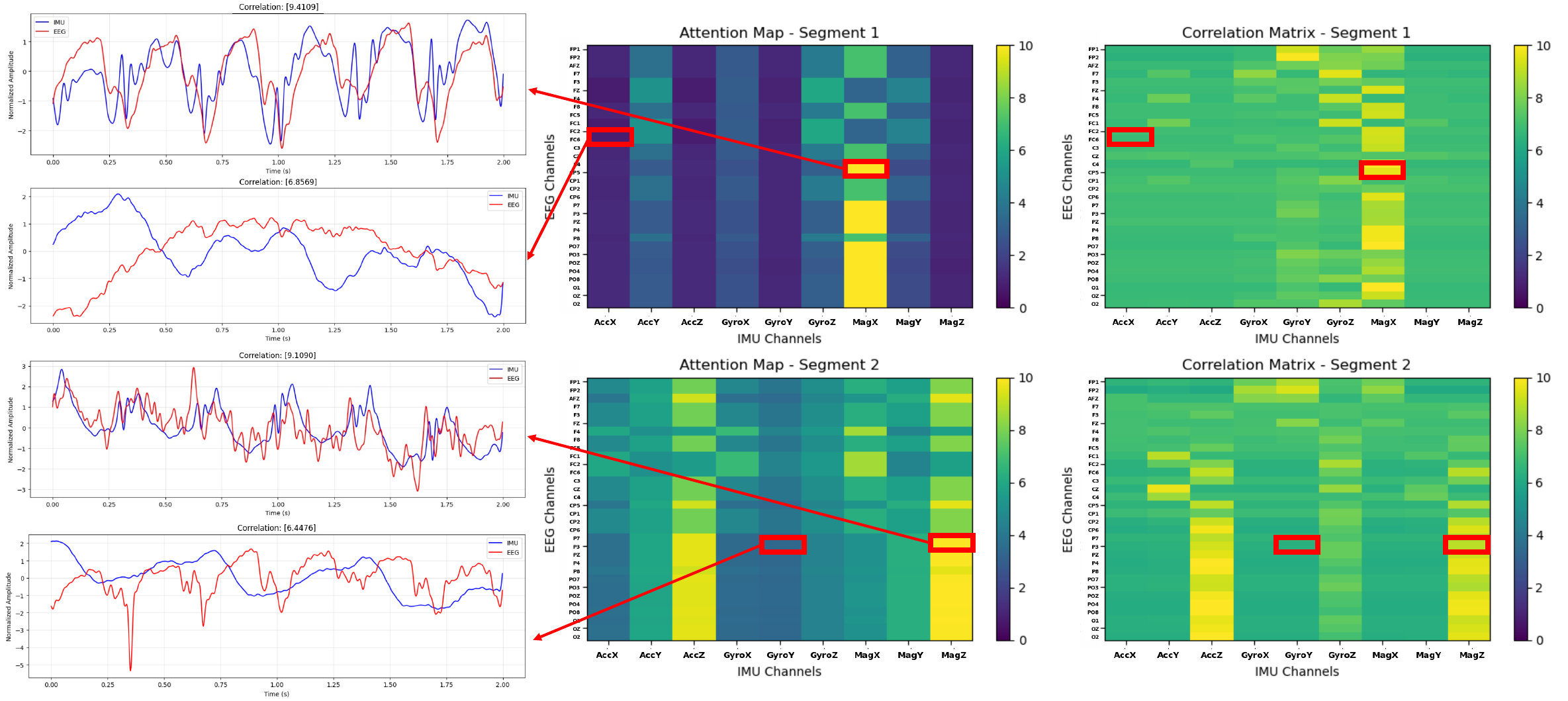}
    \caption{
    Visualization of EEG–IMU attention maps and their correspondence with correlation matrices across two 2-s signal segments. The left side shows each segment's most and least correlated EEG–IMU channel pairs, along with their time-domain waveforms and correlation values. The center panels display the learned attention maps, while the right panels show the reference correlation matrices computed via frequency-domain analysis. Red boxes highlight the same EEG–IMU pairs across all three views. The close correspondence between attention weights and true correlations indicates that the model effectively learns to identify and emphasize motion-related dependencies in the EEG signal.
    }
    \label{fig3}
\end{figure*}

LaBraM is a transformer-based neural architecture built to learn versatile EEG representations from large-scale, heterogeneous datasets \cite{jiang2024large}. It processes EEG signals by dividing them into channel-time segments and encoding them with a neural tokenizer trained to extract key spectral characteristics. By leveraging over 2,500 hours of EEG recordings from various domains, LaBraM builds rich spatiotemporal features that adapt well to a wide range of downstream BCI tasks \cite{jiang2024large}.

In our work, we use the base version of LaBraM, {\ttfamily vqnsp\_encoder\_base\_decoder\_3x200x12}, to encode motion-contaminated EEG segments into a low-dimensional representation of size 64. Each input frame contains 1 second of EEG data sampled at 200 Hz across 32 channels, resulting in an input matrix of size $[32 \times 200]$. The temporal encoder captures detailed temporal dynamics using 1D convolutional blocks. These patch embeddings are further enriched with learnable spatial and temporal embeddings and fed into a transformer encoder. A vector-quantized tokenizer then discretizes the representations by matching them to a learned codebook trained to reconstruct the frequency-domain amplitude and phase of the EEG.

An alignment layer with two linear transformations, Layer Normalization, and a GELU activation introduces nonlinearity and maps features into a 64-dimensional latent space, with a residual connection to preserve input information. Additional linear projection layers offer flexibility for fine-tuning, as illustrated in Fig.~\ref{fig2}.

Similarly, we project the 9-axis IMU signals into the same 64-dimensional feature space using a three-layer 1D convolutional encoder. The first layer uses a kernel size of 5 to map the input to 64 feature maps, followed by MaxPool1d to reduce the temporal resolution from 200 to 100 time points. The second and third layers maintain 64 channels and apply AvgPool1d to reduce the time dimension to 50. BatchNorm1d and ReLU follow each convolutional layer to improve training stability and support nonlinear feature learning. The resulting $[64 \times 50]$ feature map is flattened and passed through a fully connected layer to produce the final 64-dimensional IMU embedding.

Next, the model uses EEG queries and IMU keys to construct an attention weight matrix that computes channel-wise weights and applies them to the IMU values to determine motion artifact contributions. This attention matrix captures the pairwise relationships between EEG and IMU channels by projecting both into a shared feature space. Notably, the learned attention heatmap closely matches the ground truth correlation matrix derived from time–frequency analysis (Fig. \ref{fig3}). This visual agreement indicates that the model successfully learns to focus on IMU channels that are truly correlated with EEG motion artifacts. One major concern was whether EEG and IMU features, originating from fundamentally different signal modalities, could be projected meaningfully into the same latent space. If not, the attention mechanism would produce uniform or trivial weights, as observed in early training. To address this, we froze the pretrained EEG (LaBraM) and IMU encoders. We introduced a supervision loss that aligns the attention scores (prior to softmax) with a scaled correlation matrix, computed by subtracting 0.5 and multiplying by 20. This setup trains only the projection layers responsible for generating attention queries and values, ensuring meaningful attention between EEG and IMU inputs and aligning both modalities within a shared representation space. 

We incorporate an Artifact Gate Layer, implemented as a multilayer perceptron (MLP), which takes EEG motion artifact as input and outputs a gating signal ranging from 0 to 1, indicating the proportion of artifact to be removed. The artifact-suppressed features are then passed through a multi-layer alignment network that projects them back into the latent codebook space. Two residual connections are incorporated within the artifact removal module to preserve essential EEG structure throughout the process.

The LaBraM decoder reconstructs EEG signals in the time domain from discrete neural tokens using LaBram quantizer layer. These tokens pass through several transformer blocks, followed by average pooling and a Time Decoder Layer. The Time Decoder Layer consists of four fully connected sublayers. The first two layers expand and process the feature representations with ReLU activations and dropout for regularization. The third layer compresses the intermediate output back to the embedding dimension using a Tanh activation, followed by a final projection that maps the features to the reconstructed EEG signal space. The base LaBraM model contains approximately 8.6 million parameters, while our fine-tuned version comprises around 9.2 million parameters. We used data from 11 participants, totaling 5 hours and 52 minutes of EEG and IMU recordings, approximately 0.2346\% of the 2500 hours used to train the base LaBraM model.

\subsection{Weighted Frequency Correlation for Evaluation }

For fine-tuning, we use Weighted Frequency Correlation as the evaluation metric to assess the effectiveness of motion artifact removal. Specifically, we transform both EEG and IMU signals into the frequency domain and compute their correlation. A lower correlation value indicates more effective suppression of motion-related artifacts in EEG when using IMU as a reference signal.

We first apply the Fast Fourier Transform (FFT) to the EEG and IMU signals to extract amplitude and phase components. We then standardize these components using a layer-wise normalization function to ensure numerical stability and consistent scaling:

\begin{equation}
    \text{std\_norm}(X) = \frac{X - \text{mean}(X)}{\text{std}(X) + \epsilon}, \quad \epsilon = 10^{-8}
\end{equation}

Next, we isolate the 0–20 Hz band, corresponding to the first 40 frequency bins (given a 100 Hz Nyquist frequency for 200 Hz sampling), where motion artifacts primarily occur.

For each 1s data frame, we calculate the Pearson correlation coefficients between EEG and IMU across amplitude and phase spectra, denoted respectively as $\rho^{amp}_{ij}(f)$ and $\rho^{phase}_{ij}(f)$ at frequency $f$. Amplitude is more strongly affected by motion artifacts than phase; therefore, we assign a weight of 0.7 to the amplitude correlation and 0.3 to the phase correlation. We then combine these into a weighted sum to reflect the substantial impact of motion on amplitude:

\begin{equation}
    \bar{\rho}^{(p)}_{ij} = \frac{1}{F} \sum_{f=1}^{F} \left[ 0.7 \cdot \rho^{amp}_{ij}(f) + 0.3 \cdot \rho^{phase}_{ij}(f) \right]
\end{equation}

Here, $F = 40$ is the number of frequency points within the 0–20 Hz range, and $i$ and $j$ index the 32 EEG and 9 IMU channels, respectively. Finally, we compute a scalar coherence score for each patch by averaging across all channel pairs:

\begin{equation}
    \text{Coherence}_p = \frac{1}{C_{EEG} \times C_{IMU}} \sum_{i=1}^{C_{EEG}} \sum_{j=1}^{C_{IMU}} \bar{\rho}^{(p)}_{ij}
\end{equation}

where $C_{EEG} = 32$ and $C_{IMU} = 9$. Lower coherence values indicate weaker frequency-domain correlation between the IMU and EEG, suggesting more effective removal of motion-related artifacts. We use this coherence score as both a quantitative evaluation metric and a training loss signal while fine-tuning our IMU-integrated denoising model.

\subsection{ICA and ASR for Benchmarking}
Independent Component Analysis (ICA) is a widely used blind source separation technique that decomposes EEG signals into statistically independent components \cite{jung2000removing}. It effectively separates stereotypical artifacts such as eye blinks, muscle activity, and line noise, enabling selective removal of non-neural components. In our implementation, we used \textit{ICLabel} for automatic component classification, removing components that exceeded the following confidence thresholds: Muscle ($>$90\%), Eye ($>$40\%), Heart ($>$90\%), Line Noise ($>$50\%), Channel Noise ($>$90\%), and Other ($>$90\%). However, ICA often requires manual or semi-automated identification of artifact components and struggles transient, high-amplitude disturbances.

Artifact Subspace Reconstruction (ASR) mitigates these limitations by automatically detecting and suppressing high-variance artifacts in real-time EEG data. It estimates an artifact threshold based on clean reference segments in the principal component space and reconstructs cleaner EEG signals accordingly. ASR’s compatibility with online applications and its ability to preserve neural activity make it an ideal benchmark for evaluating modern artifact removal methods \cite{chang2019evaluation}. We use these two algorithms as baselines to benchmark the performance of our large-model-based fine-tuning pipeline.

\section{Result Evaluation}
\begin{table*}[t]
\centering
\caption{Comparison of methods for IMU–EEG correlation}
\begin{tabular}{@{}lcccc@{}}
\toprule
\textbf{Condition} & \textbf{Time Scale} & \textbf{Raw EEG} & \textbf{ASR + ICA} & \textbf{LaBraM Fine-tune (our)} \\
\midrule
\multirow{3}{*}{Slight Running (2.0 m/s), ses-05} 
& 10 s  & $0.5309 \pm 0.0619$ & $0.3006 \pm 0.0606$ & $\mathbf{0.2560 \pm 0.0122}$ \\
& 30 s  & $0.5354 \pm 0.0678$ & $0.3201 \pm 0.0616$ & $\mathbf{0.2591 \pm 0.0097}$ \\
& 1 min & $0.5327 \pm 0.0616$ & $0.3334 \pm 0.0437$ & $\mathbf{0.2602 \pm 0.0096}$ \\
\midrule
\multirow{3}{*}{Fast Walking (1.6 m/s), ses-04} 
& 10 s  & $0.5548 \pm 0.0354$ & $0.3173 \pm 0.0642$ & $\mathbf{0.2344 \pm 0.0069}$ \\
& 30 s  & $0.5255 \pm 0.0547$ & $0.3030 \pm 0.0693$ & $\mathbf{0.2574 \pm 0.0093}$ \\
& 1 min & $0.5535 \pm 0.0380$ & $0.3015 \pm 0.0598$ & $\mathbf{0.2511 \pm 0.0084}$ \\
\midrule
\multirow{3}{*}{Slow Walking (0.8 m/s), ses-03} 
& 10 s  & $0.5438 \pm 0.0320$ & $0.2907 \pm 0.0549$ & $\mathbf{0.2511 \pm 0.0099}$ \\
& 30 s  & $0.5566 \pm 0.0539$ & $0.3039 \pm 0.0827$ & $\mathbf{0.2506 \pm 0.0089}$ \\
& 1 min & $0.5605 \pm 0.0460$ & $0.3064 \pm 0.0624$ & $\mathbf{0.2698 \pm 0.0101}$ \\
\bottomrule
\end{tabular}
\label{tab:artifact_comparison}
\end{table*}

\subsection{EEG–IMU Attention Mapping Analysis}

\begin{figure}[htbp]
    \centering
    \includegraphics[width=\linewidth]{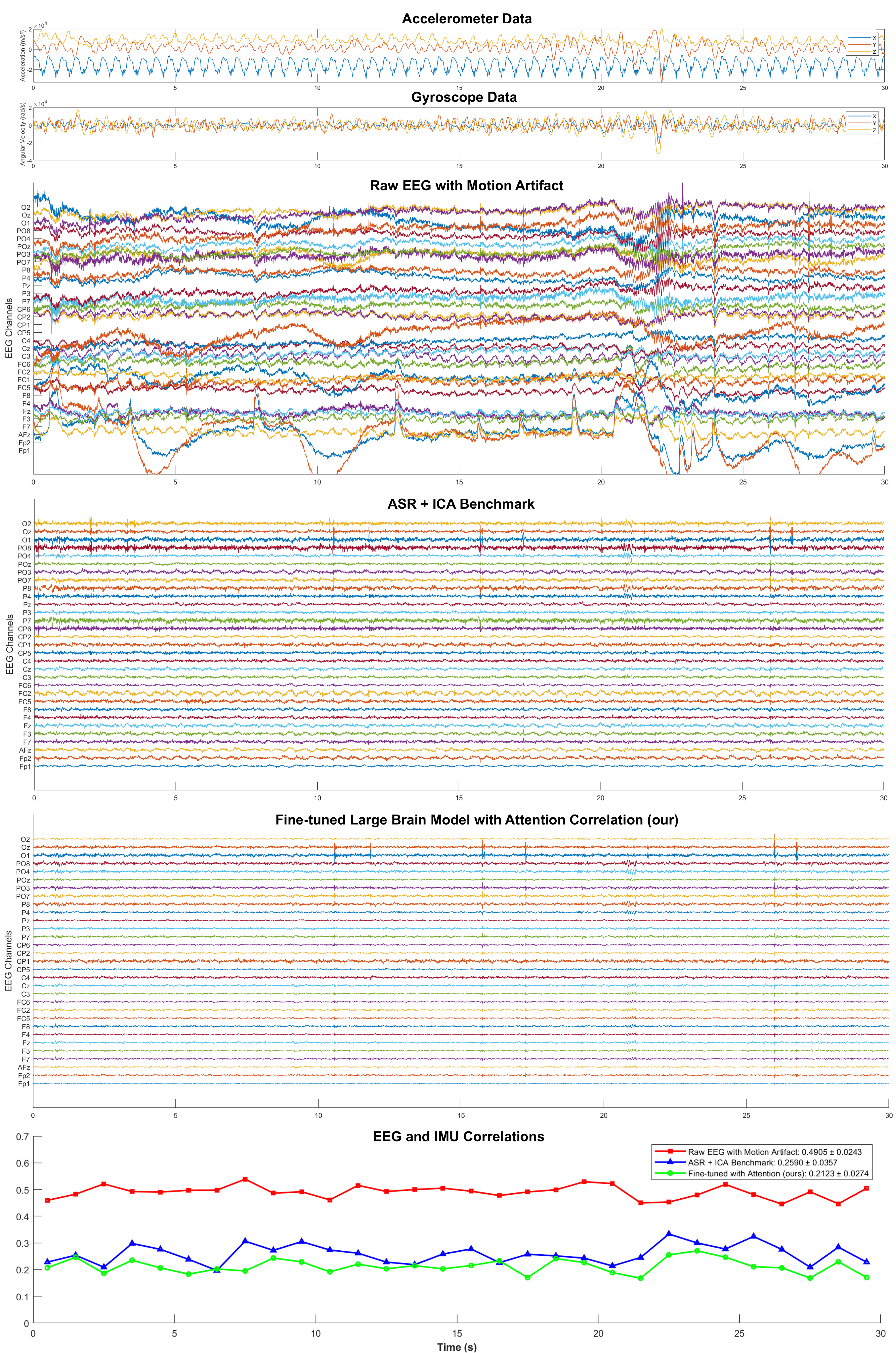}
    \caption{
   EEG denoising comparison during fast walking (1.6 m/s) over a 30-second interval. Top panels display the raw IMU signals from the accelerometer and gyroscope. The third panel presents the raw EEG with strong motion artifacts, followed by the ASR + ICA benchmark output and the result from our fine-tuned Large Brain Model using attention-based IMU integration. The bottom plot illustrates the EEG–IMU correlation over time for each method. Our model achieves the lowest average correlation (0.2123 ± 0.0274), indicating superior motion artifact suppression compared to both raw EEG (0.4905 ± 0.0243) and ASR + ICA (0.2590 ± 0.0357).
    }
    \label{fig4}
\end{figure}

Figure~\ref{fig3} presents a comparative analysis of the learned EEG–IMU attention maps and the corresponding frequency-domain correlation matrices across two representative signal segments. For each segment, we highlight the most and least correlated EEG–IMU channel pairs. The temporal waveforms (left) depict signal alignment, while the attention maps (center) and correlation matrices (right) visualize the corresponding weight distributions across all channel combinations.

The learned attention weights closely mirror the reference correlation patterns, showing clear structural similarity. In both segments, the attention mechanism accurately assigns higher weights to IMU channels with stronger correlation to EEG signals and lower weights to those with minimal true correlations. This consistency indicates that the model successfully captures meaningful cross-modal dependencies by aligning EEG and IMU features in a shared latent space.

Moreover, the model’s ability to distinguish high- from low-correlation pairs without explicit supervision during inference underscores the effectiveness of the attention-based projection mechanism. These results empirically validate the representational alignment across modalities and confirm the attention module's capability to serve as a reliable estimator of motion-related interference in EEG signals.

\subsection{EEG IMU Correlation Analysis}

Table~\ref{tab:artifact_comparison} compares the IMU–EEG correlation under three different motion conditions (slight running, fast walking, and slow walking) and three time scales (10 seconds, 30 seconds, and 1 minute). Across all conditions and durations, the raw EEG signals show the highest correlation with IMU signals, indicating the strong presence of motion artifacts. Applying conventional artifact removal methods, such as ASR combined with ICA,  noticeably reduces the correlation, indicating partial suppression of motion-induced noise.

In contrast, our fine-tuned LaBraM model consistently achieves the lowest correlation scores across all scenarios. For instance, under slight running conditions, the correlation drops from 0.5309 (raw) and 0.3006 (ASR+ICA) to 0.2560 using our method on the 10-second scale. We observed similar improvements in both fast and slow walking conditions, with reductions exceeding 50\% compared to raw EEG and over 13\% compared to ASR+ICA. These results demonstrate that our approach more effectively disentangles motion artifacts from neural signals, even under intense movement and across multiple timescales.

\subsection{Comparison of Results Across Time Scales and Motion Tasks}

We evaluate our proposed method across three time scales (10 seconds, 30 seconds, and 1 minute) and three motion-related tasks—slight running (2.0 m/s), fast walking (1.6 m/s), and slow walking (0.8 m/s)—benchmarking against the widely adopted ASR + ICA pipeline. As illustrated in Fig.~\ref{fig2}, our model processes 1-second EEG segments sequentially and outputs a corresponding correlation loss per second. Fig.~\ref{fig4} presents the comparative results for the 30-second fast walking condition. Raw EEG signals show prominent motion artifacts across all conditions, particularly during high-intensity movements such as running. While ASR + ICA significantly attenuates some of these fluctuations, residual artifacts and temporal discontinuities remain visible, especially in the midline and frontal regions. 
In contrast, our method maintains the lowest correlation levels throughout the entire interval.
Additional evaluation figures for all time scales and motion conditions appear in the Appendix.

\section{Discussion and Conclusion}

In this study, we propose a novel framework that fine-tunes a Large Brain Model (LaBraM) for the downstream task of EEG motion artifact removal using multimodal, attention-based integration with IMU signals. This approach addresses a critical methodological gap, as most prior techniques rely on single-modality processing and conventional filters, such as adaptive digital filters, ICA, and ASR. By leveraging pretrained spatiotemporal representations and introducing an attention mechanism that aligns EEG and IMU features within a shared latent space, our method achieves superior artifact suppression across various motion intensities and time scales. The attention weights not only guide effective denoising but also correlate strongly with frequency-domain correlation measures, offering an interpretable and physiologically meaningful explanation of cross-modal dependencies.

Despite these promising results, the current evaluation remains limited to a subset of motion conditions and participants. Due to time constraints, we have not yet compared our method with other state-of-the-art approaches, which we plan to pursue in future work. We also need additional efforts to validate the model’s generalizability across more diverse datasets, including those with different sensor configurations, movement profiles, and subject variability. Expanding the dataset and conducting extensive cross-subject and cross-session evaluations will be essential for ensuring robustness in real-world applications. Nonetheless, this work lays a foundational step toward scalable and explainable multimodal artifact removal in mobile EEG, bridging the gap between large deep learning models and practical brain–computer interface deployment.



\bibliographystyle{ieeetr}
\bibliography{references}

 


\clearpage 

\twocolumn[{
  \begin{center}
    \section*{Appendix}
    \vspace{0.5em}
  \end{center}
  \renewcommand{\thefigure}{\arabic{figure}} 
  \setcounter{figure}{0}
  \begin{center}
    \includegraphics[width=0.8\linewidth]{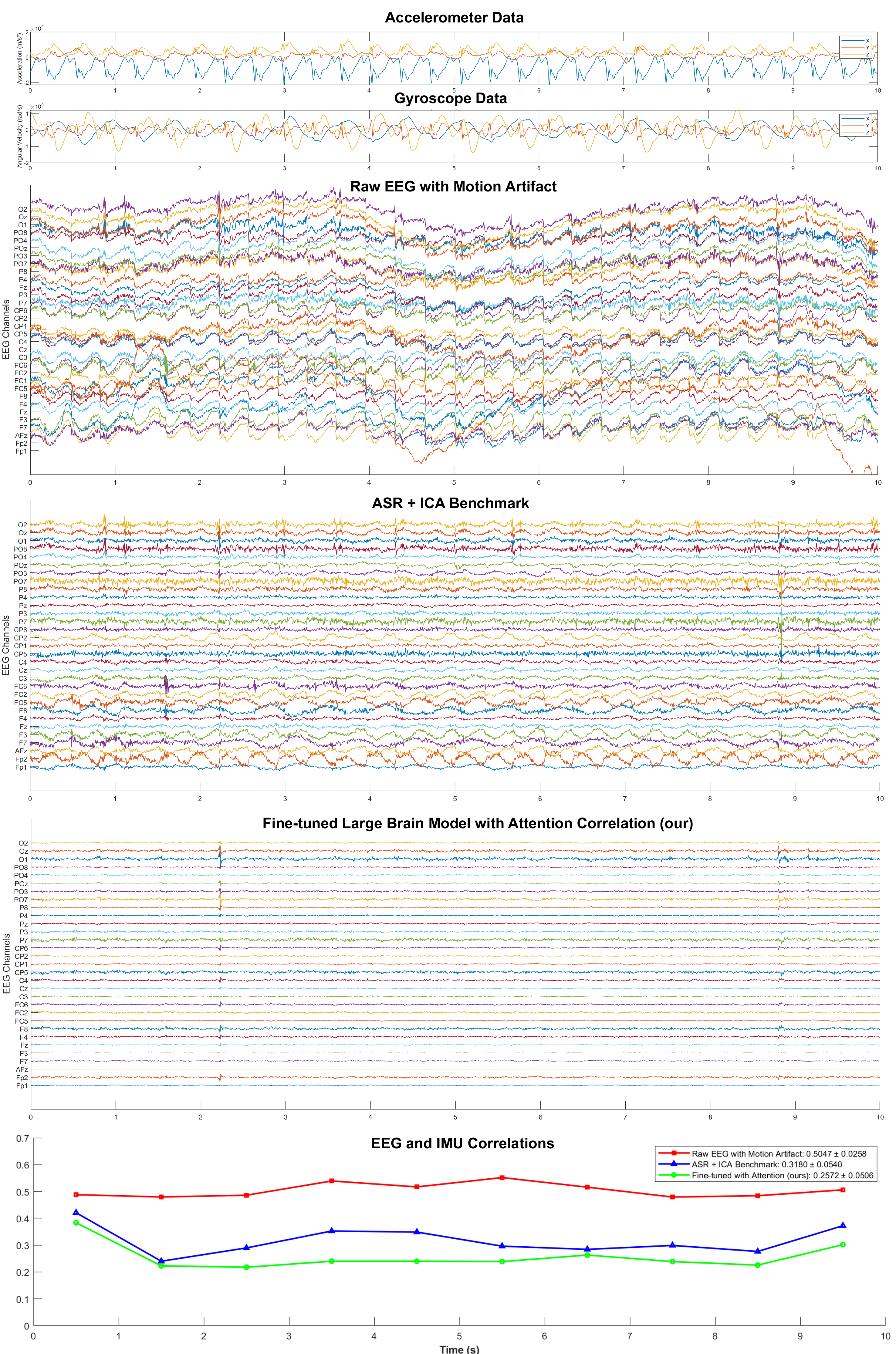}
    \captionof{figure}{Slight Running (2.0 m/s), 10-second window.}
  \end{center}}
]

\begin{figure*}[htbp]
    \centering
    \includegraphics[width=0.8\linewidth]{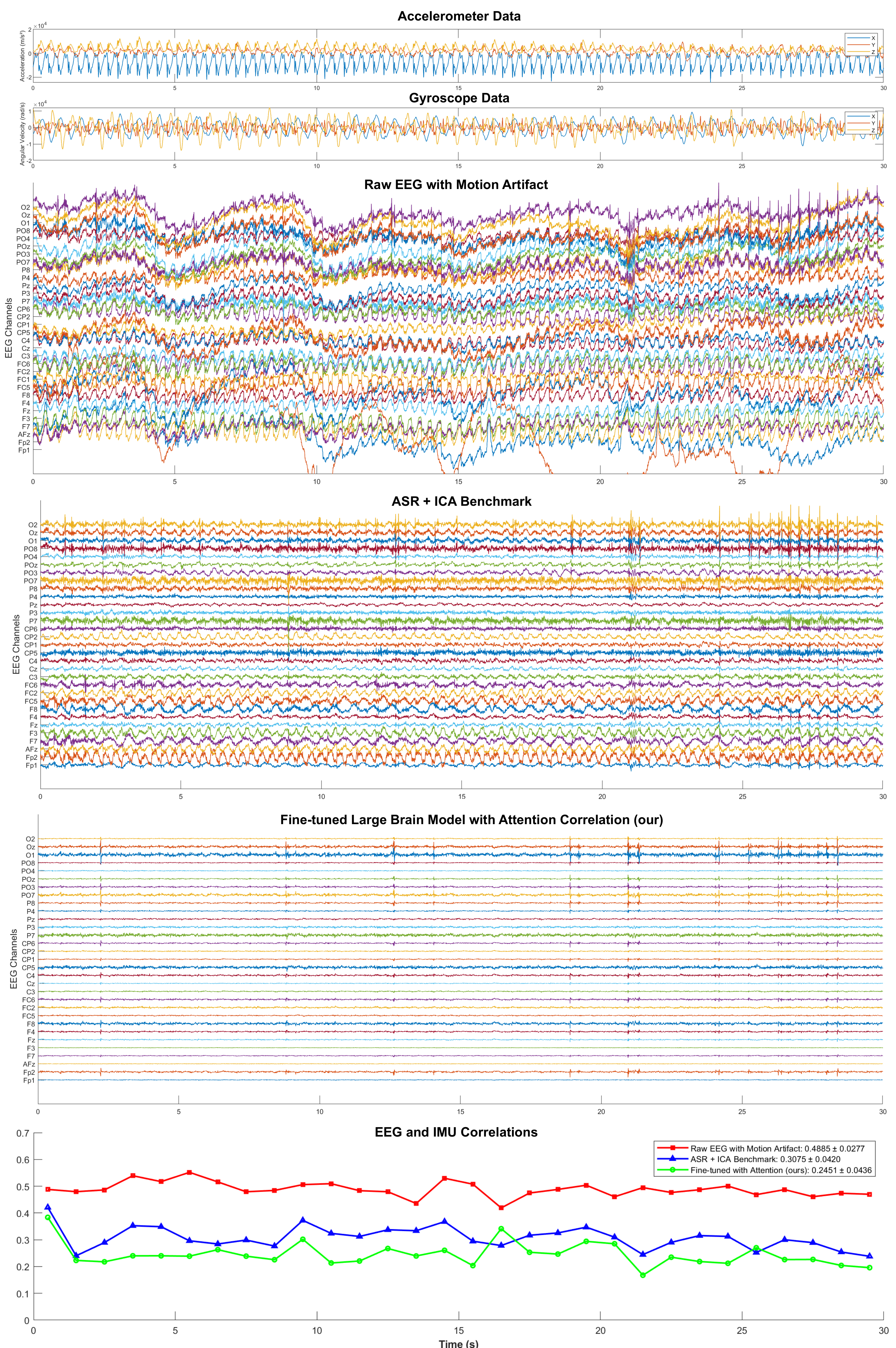}
    \caption{Slight Running (2.0 m/s), 30-second window.}
\end{figure*}

\begin{figure*}[htbp]
    \centering
    \includegraphics[width=0.8\linewidth]{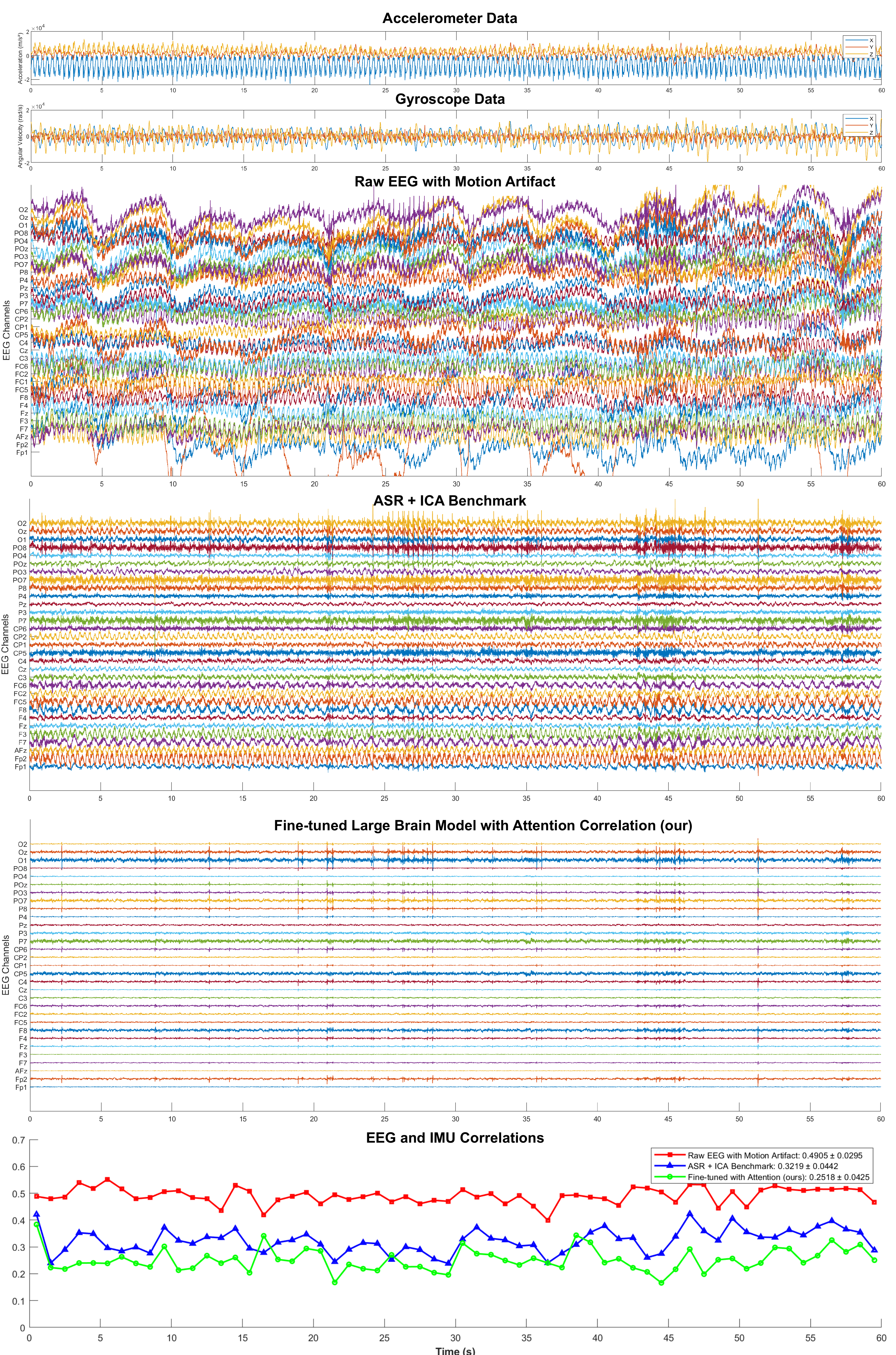}
    \caption{Slight Running (2.0 m/s), 60-second window.}
\end{figure*}

\begin{figure*}[htbp]
    \centering
    \includegraphics[width=0.8\linewidth]{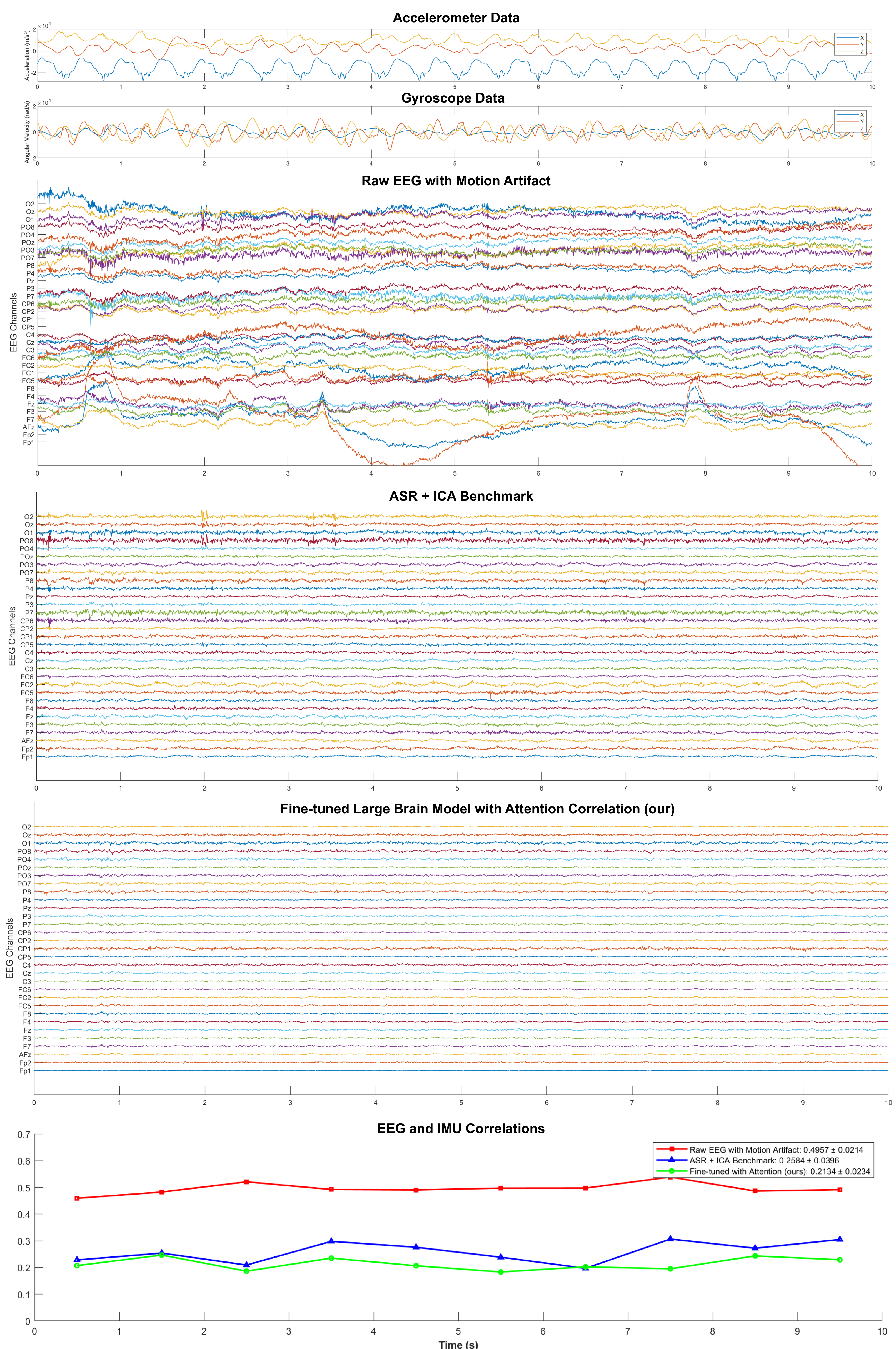}
    \caption{Fast Walking (1.6 m/s), 10-second window.}
\end{figure*}

\begin{figure*}[htbp]
    \centering
    \includegraphics[width=0.8\linewidth]{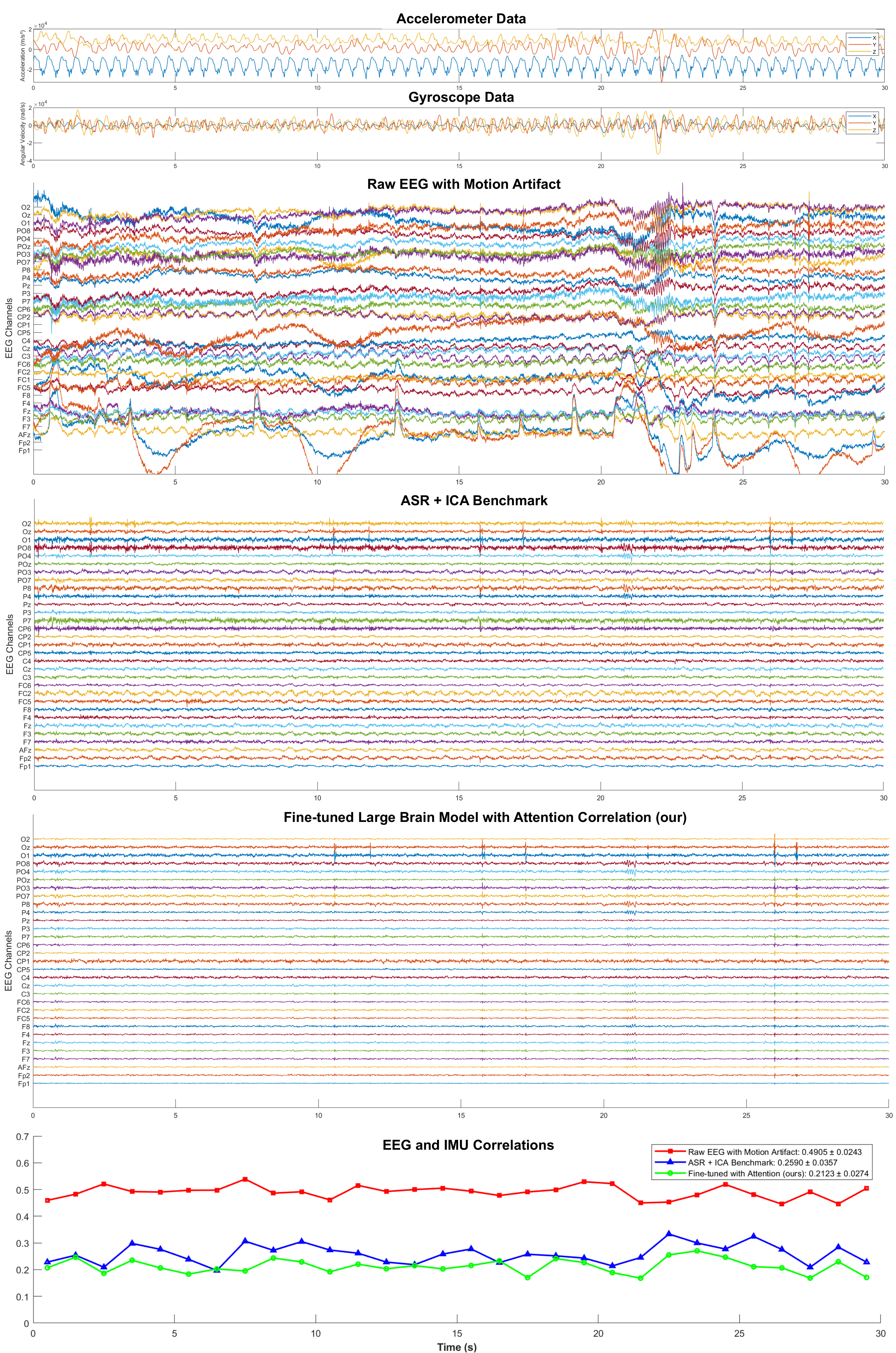}
    \caption{Fast Walking (1.6 m/s), 30-second window.}
\end{figure*}

\begin{figure*}[htbp]
    \centering
    \includegraphics[width=0.8\linewidth]{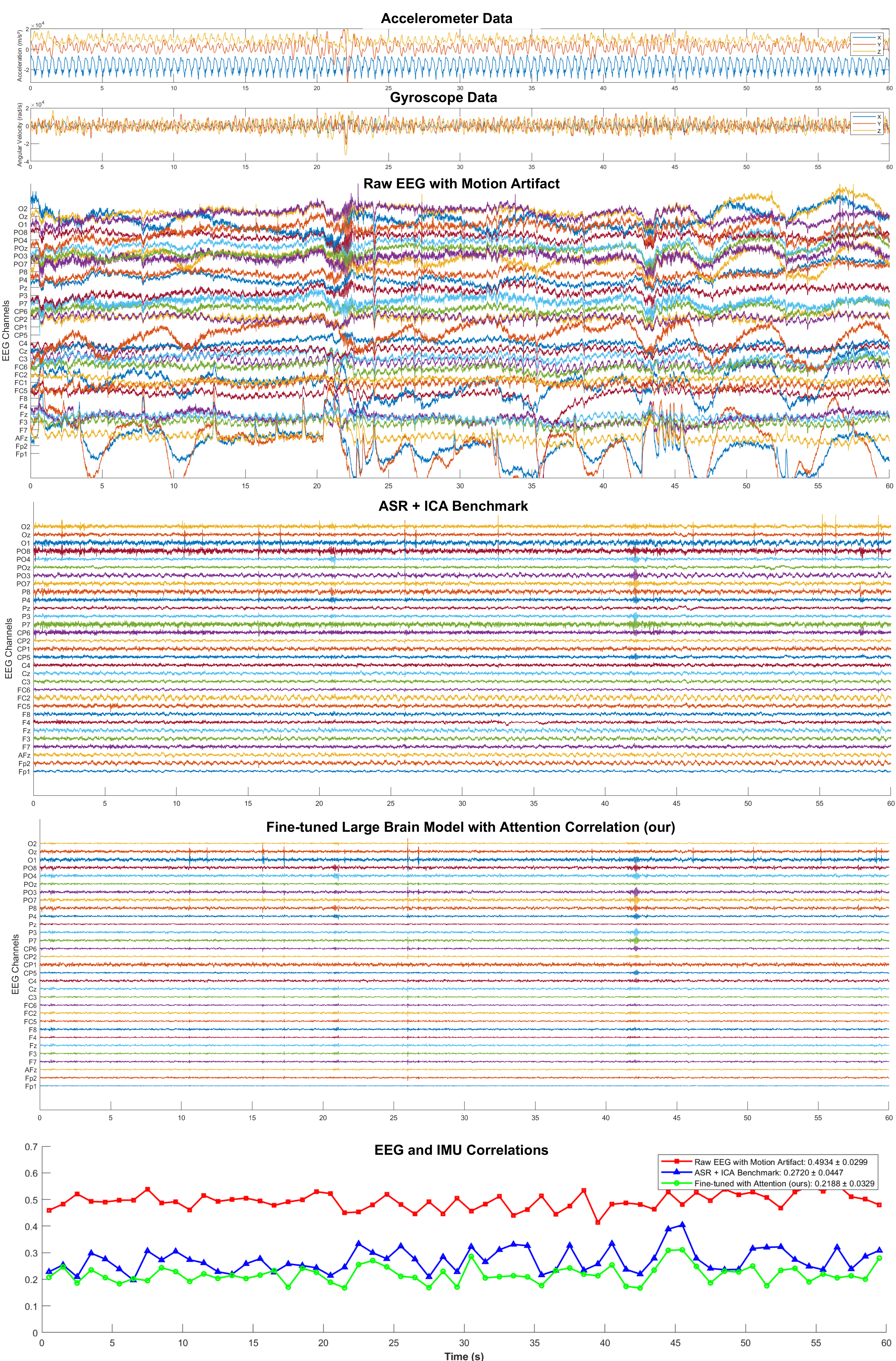}
    \caption{Fast Walking (1.6 m/s), 60-second window.}
\end{figure*}

\begin{figure*}[htbp]
    \centering
    \includegraphics[width=0.8\linewidth]{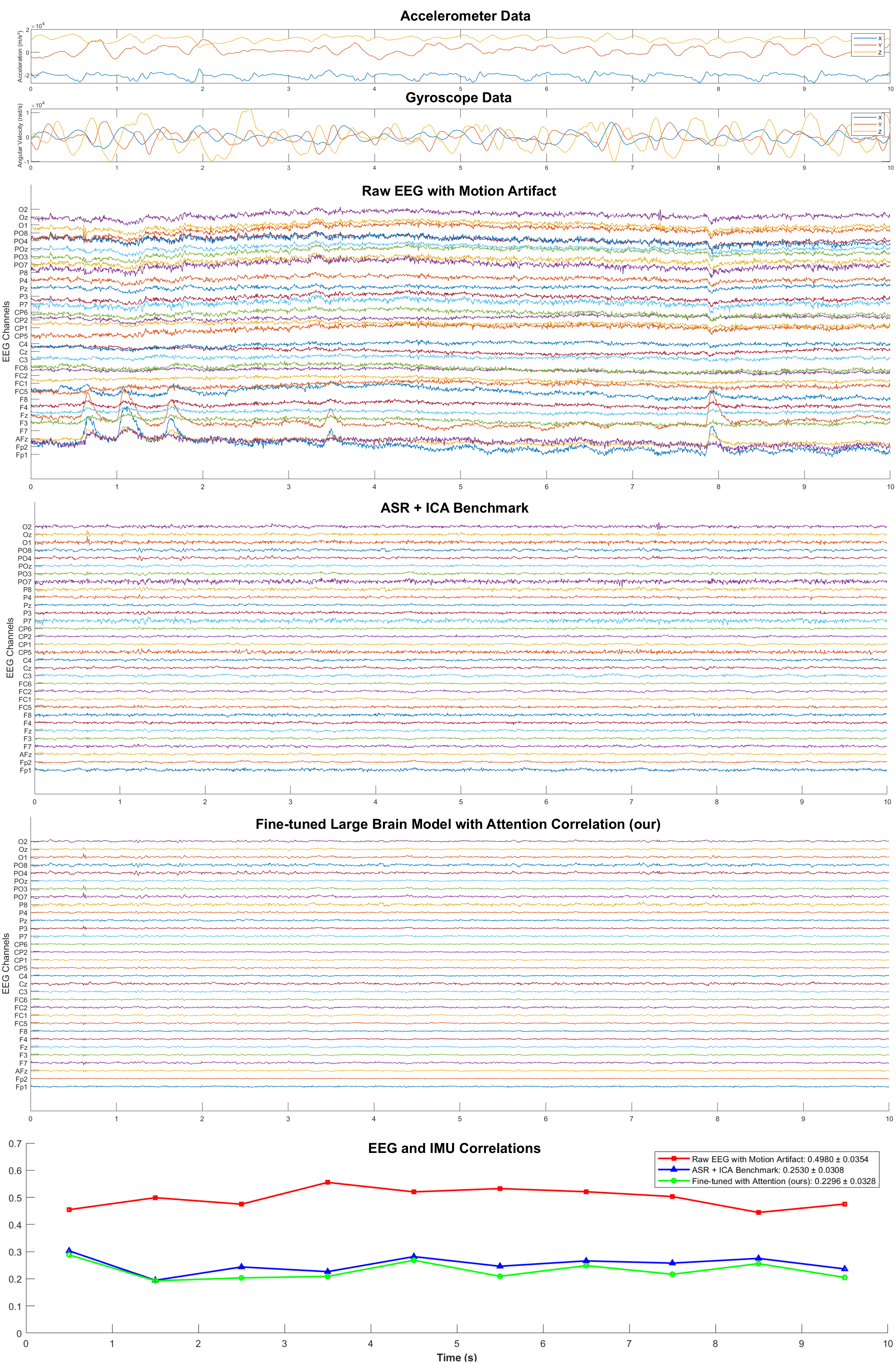}
    \caption{Slow Walking (0.8 m/s), 10-second window.}
\end{figure*}

\begin{figure*}[htbp]
    \centering
    \includegraphics[width=0.8\linewidth]{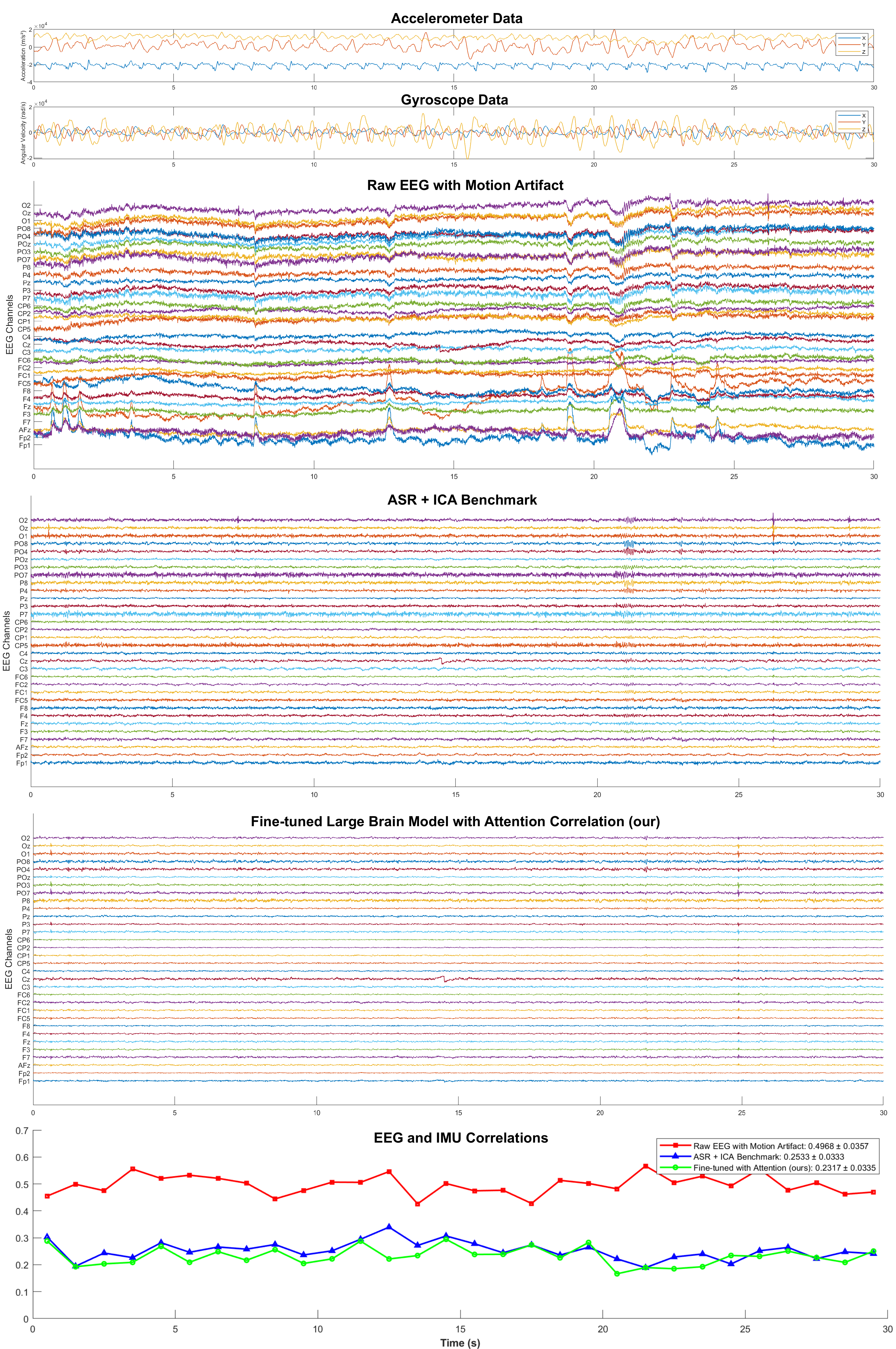}
    \caption{Slow Walking (0.8 m/s), 30-second window.}
\end{figure*}

\begin{figure*}[htbp]
    \centering
    \includegraphics[width=0.8\linewidth]{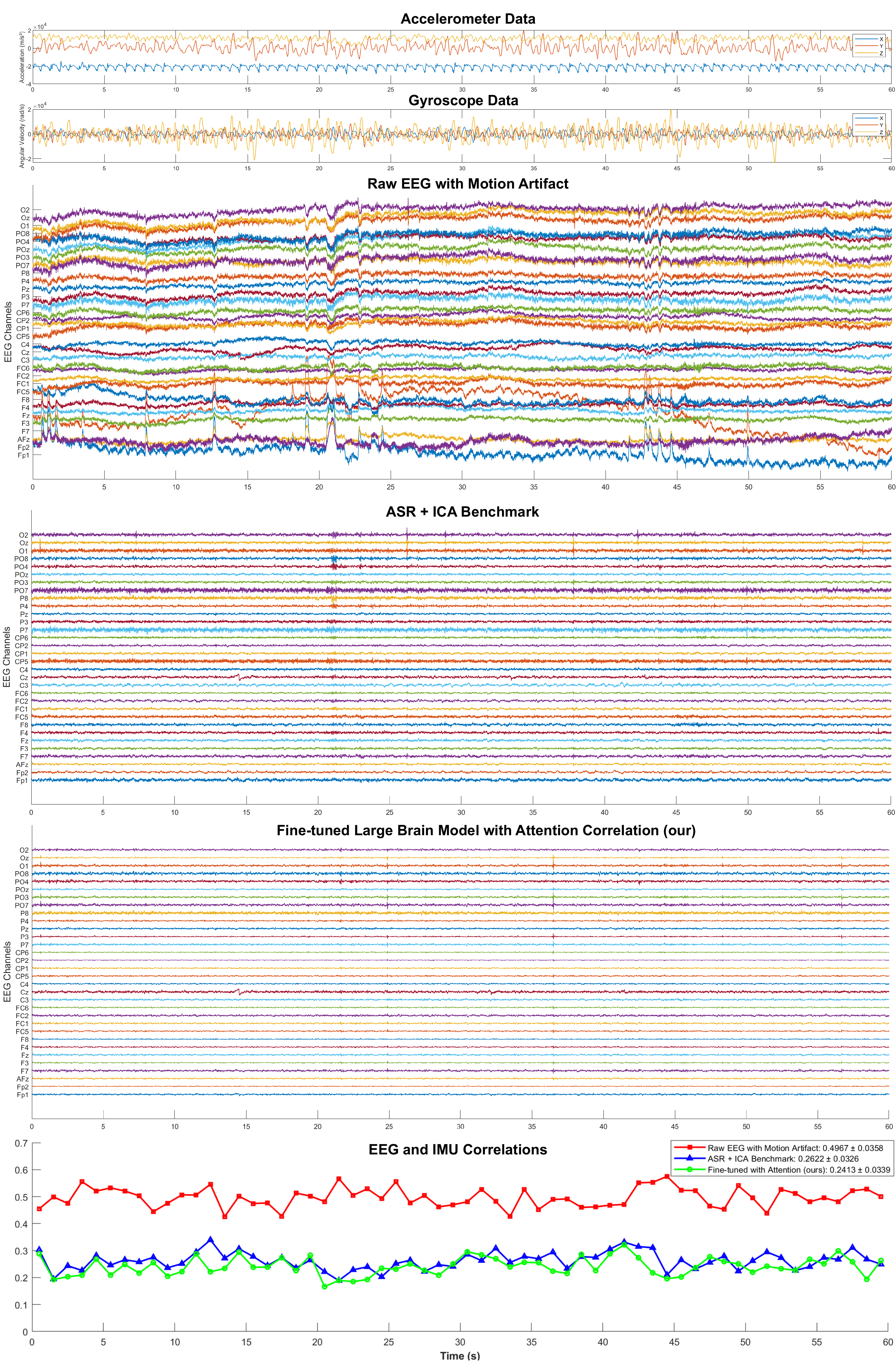}
    \caption{Slow Walking (0.8 m/s), 60-second window.}
\end{figure*}

\begin{figure*}[t]
    \centering
    \includegraphics[width=0.85\linewidth]{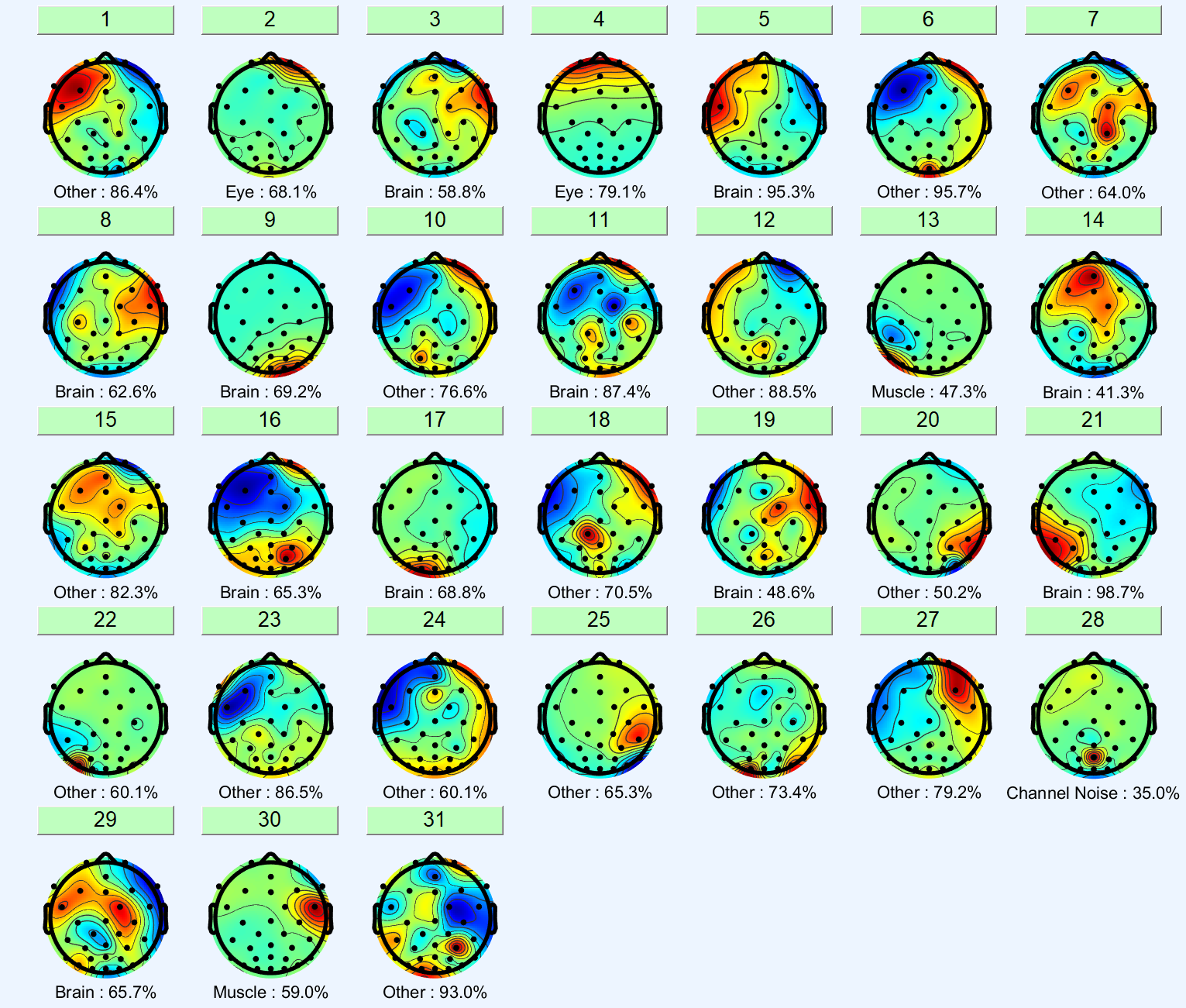}
    \caption{ICLabel-based classification of ICA components during slight running.}
\end{figure*}

\end{document}